\documentclass[sigconf]{acmart}
\AtBeginDocument{%
  \providecommand\BibTeX{{%
    \normalfont B\kern-0.5em{\scshape i\kern-0.25em b}\kern-0.8em\TeX}}}

\setcopyright{acmcopyright}
\copyrightyear{.}
\acmYear{.}
\acmDOI{.}

\acmConference[]{}{}{}
\acmBooktitle{}
\acmPrice{.}

\usepackage{latexsym}
\usepackage{arydshln}
\usepackage{caption}
\usepackage{subcaption}
\usepackage{tcolorbox}

\usepackage{graphicx}
\usepackage{hhline}
\usepackage{bm}
\usepackage{natbib}
\usepackage{bbm}
\usepackage{amsmath}
\usepackage{amsthm}
\usepackage[ruled,linesnumbered]{algorithm2e}

\let\oldnl\nl
\newcommand{\nonl}{\renewcommand{\nl}{\let\nl\oldnl}}
\newcommand\ci{\perp\!\!\!\perp}

\makeatletter
\newtheorem*{rep@theorem}{\rep@title}
\newcommand{\newreptheorem}[2]{%
\newenvironment{rep#1}[1]{%
 \def\rep@title{#2 \ref{##1}}%
 \begin{rep@theorem}}%
 {\end{rep@theorem}}}
\makeatother

\newtheorem{assumptions1}{Assumption}
\newtheorem{theorem1}{Theorem}
\newtheorem{lemma1}{Lemma}
\newtheorem{proposition1}{Proposition}
\newreptheorem{theorem}{Theorem}
\newreptheorem{lemma}{Lemma}
\newreptheorem{proposition}{Proposition}

\begin{document}

\title{Synthesized Difference in Differences}

\author{Eric V. Strobl}
\author{Thomas A. Lasko}
\affiliation{%
  \institution{Vanderbilt University Medical Center}
  \city{Nashville}
  \state{Tennessee}
  \country{USA}
}
\email{}
\email{}

\renewcommand{\shortauthors}{Strobl and Lasko}

\begin{abstract}
We consider estimating the conditional average treatment effect for everyone by eliminating confounding and selection bias. Unfortunately, randomized clinical trials (RCTs) eliminate confounding but impose \textit{strict} exclusion criteria that prevent sampling of the entire clinical population. Observational datasets are more inclusive but suffer from confounding. We therefore analyze RCT and observational data simultaneously in order to extract the strengths of each. Our solution builds upon Difference in Differences (DD), an algorithm that eliminates confounding from observational data by comparing outcomes before and after treatment administration. DD requires a parallel slopes assumption that may not apply in practice when confounding shifts across time. We instead propose Synthesized Difference in Differences (SDD) that infers the correct (possibly non-parallel) slopes by linearly adjusting a conditional version of DD using additional RCT data. The algorithm achieves state of the art performance across multiple synthetic and real datasets even when the RCT excludes the majority of patients. 
\end{abstract}





\maketitle

\section{Introduction} \label{sec_intro}
Scientists often use randomized clinical trials (RCTs) to infer causal relations. RCTs offer strong \textit{internal validity} by eliminating confounding with randomized \textit{treatment assignment} $T$, which we assume to be binary. This allows investigators to estimate the causal effect of treatment on an outcome given patient covariates. Randomization however can be difficult, unethical, time consuming or expensive to perform. The majority of RCTs therefore contain many exclusion criteria that sacrifice \textit{external validity} by limiting who can receive treatment. Standard machine learning algorithms trained on RCT data can in turn fail to accurately estimate the treatment effect across the entire clinical population.

Observational datasets on the other hand employ few exclusion criteria. This allows investigators to fit complex non-linear models across the broader population. Observational datasets are however prone to confounding because common causes may dictate both treatment assignment and outcome. As a result, the estimated model may not match the true treatment effect no matter how complicated the fit. Observational datasets thus have strong external but weak internal validity, precisely the opposite situation encountered with RCTs.

The above issues motivate us to develop an algorithm that addresses the following problem:
\begin{tcolorbox}
\textbf{Objective.} Estimate the treatment effect across the entire clinical population by analyzing observational and RCT data simultaneously in order to achieve strong external and internal validity.
\end{tcolorbox}
\noindent The proposed solution builds upon \textit{Difference in Differences} (DD), a method frequently used in econometrics to estimate the treatment effect using additional time information \citep{Card93}. The algorithm assumes that the observational data contains measurements of the outcome both before and after treatment administration. The difference between the two treatment groups \textit{before} treatment administration represents the confounding effect, so DD subtracts out this quantity \textit{after} treatment administration in order to recover the treatment effect. DD therefore assumes that the confounding effect remains constant across time. We provide further details in Section 4.

The confounding effect may unfortunately differ between the two time points in practice. For example, physicians may prescribe the active treatment to the sickest patients who naturally worsen without medical intervention. The healthiest patients on the other hand may recover even with placebo, so that the confounding effect increases over time. In this paper, we modify DD to account for time varying confounding effects by leveraging additional RCT data.

\textit{Cross-design synthesis} refers to a broader collection of methods that leverage RCT and observational data simultaneously in order to strengthen both external and internal validity \citep{Droitcour93}. While our work is the first to synthesize DD, a number of other methods also attempt to eliminate confounding using both data types without time information. Most methods however assume that the confounding effect can be represented as a linear transformation of a predefined set of basis functions (e.g., \citep{Jackson17,Kallus18}). We in contrast only assume that the \textit{difference} between the pre- and post-treatment confounding effects decomposes linearly. The confounding effect may therefore take on a highly non-linear form within each time step. 

We organize the rest of this paper as follows. We first introduce the potential outcomes framework in Section \ref{sec_PO}. We then describe related work in Section \ref{sec_RW}. Section \ref{sec_DD} details DD and then extends it to the conditional setting. We introduce the proposed Synthesized Difference in Differences (SDD) algorithm in Section \ref{sec_SDD}. Experimental results in Section \ref{sec_exp} highlight the superiority of SDD in estimating the treatment effect from both synthetic and real data. We finally conclude the paper in Section \ref{sec_concl}. Most proofs are located in the Appendix.

\section{Potential Outcomes} \label{sec_PO}
We adopt the potential outcomes framework, where we posit the existence of potential outcomes $Y(T)$ for $T=0$ and $T=1$. However, we can only observe one potential outcome $Y(t)$ for the subject assigned to $T=t$ in practice. 

Physicians often assign treatments based on patient covariates $\bm{X}$. In general, $T$ and $Y(T)$ are dependent given $\bm{X}$ in the distribution of the observational population. For example, a physician may prescribe mirtazapine to depressed patients experiencing insomnia and decreased appetite. Patients with those symptoms in turn experience a greater reduction in depression compared to patients without those symptoms. The covariates $\bm{X}$ can therefore confound treatment assignment and outcome.

RCTs fortunately overcome confounding by randomizing treatment assignment. Randomization over the \textit{entire} clinical population ensures that assignment and potential outcomes are independent given $\bm{X}$, denoted by $\{Y(0),Y(1)\} \ci T | \bm{X}$. Assignment therefore occurs without predicting outcome a priori, i.e. $\mathbb{P}(Y(T)|\bm{X}) = \mathbb{P}(Y(T)|T,\bm{X})$. In the mirtazapine example, a physician already knows that patients who experience insomnia and decreased appetite respond better to mirtazapine than those who do not because of the drug's actions on histamine and serotonin. Randomization breaks this dependency by assigning treatments regardless of suspected differential effect.

Some patients may unfortunately find randomization disturbing because they cannot ensure optimal treatment assignment. Physicians may also feel uncomfortable randomizing treatments to severely ill patients. RCTs therefore require strict \textit{exclusion criteria} in practice, or rules prohibiting recruitment into the trial before assignment.\footnote{We can convert any \textit{inclusion criterion} to an exclusionary one by negation.} Exclusion criteria limit sample size and impose \textit{selection bias} according to covariates $\bm{S}$ because investigators can only measure treatment effect for a subset of the clinical population. Without loss of generality, let $\bm{S}$ denote a binary random variable equal to zero when a patient meets exclusion criteria and one otherwise. RCTs therefore sample covariates from the distribution $\mathbb{P}(\bm{X}| \bm{S}=1)$ whose support $\mathcal{S}_R$ may differ from $\mathcal{S}_O$, the support of $\mathbb{P}(\bm{X})$ for the observational data. The exclusion criteria in particular imply:
\begin{assumptions1} \label{as_support}
We have $\mathcal{S}_R \subseteq \mathcal{S}_O$.
\end{assumptions1}
\noindent Any statistical procedure that makes inferences using the RCT must therefore \textit{extrapolate} to $\mathcal{S}_O \setminus \mathcal{S}_R$ in order make inferences about the entire clinical population.

While RCTs suffer from selection bias, they still randomize among the recruited to eliminate confounding therein so that:
\begin{assumptions1}  (Unconfoundedness) \label{ax_RCT}
We have $\{Y(0),Y(1)\} \ci T | \{\bm{X},\bm{S}=1\}$ in the RCT distribution.
\end{assumptions1} 
\noindent We thus always have $\mathbb{P}(Y(T)|\bm{X},\bm{S}=1) = \mathbb{P}(Y(T)|\bm{X},T,\bm{S}=1)$ in the RCT. This equality does not necessarily hold with the observational distribution. We also require:
\begin{assumptions1} \label{as_SB}
$\{Y(0),Y(1)\} \ci \bm{S} | \bm{X}$ in the RCT distribution.
\end{assumptions1}
\noindent so that the potential outcomes only depend on $\bm{X}$ in the RCT. This is a reasonable assumption in practice because the exclusion criteria are usually known, so we can include the direct causes of $\bm{S}$ in $\bm{X}$.  

In summary, RCTs impose selection bias but eliminate confounding, while observational datasets impose confounding but eliminate selection bias. In this paper, we take advantage of the observational and RCT data \textit{simultaneously} in order to eliminate the weaknesses of each. We specifically do so in order to estimate the conditional average treatment effect (CATE) $\mathbb{E}(Y(1)-Y(0)|\bm{X}) \triangleq f(\bm{X})$ because treatment effects may differ depending on patient characteristics, so estimating the CATE helps customize treatment in the spirit of precision medicine. We would like to estimate $f(\bm{X})$ over the entire clinical population, or over $\mathcal{S}_O$, but selection bias in the RCT only allows us to sample on $\mathcal{S}_R$. We therefore seek to extrapolate the CATE from $\mathcal{S}_R$ to $\mathcal{S}_O \setminus \mathcal{S}_R$.

\section{Related Work} \label{sec_RW}
Investigators have proposed several other algorithms that generalize the CATE to $\mathcal{S}_O$. The first group of methods assume that the observational data is unconfounded with enough measured variables \citep{Rubin78,Rosenbaum83,Pearl09}. Authors therefore propose algorithms that efficiently handle high dimensions in order to estimate $f(\bm{X})$. The weakness of these methods of course lies in the unconfoundedness assumption, which we can neither guarantee nor test for in practice.

Another line of methods called inverse probability weighted estimators (IPWEs) weigh the RCT samples according to the inverse of $\mathbb{P}(\bm{S}=1|T,\bm{X})$ by applying the exclusion criteria to the observational data \citep{Rosenbaum83,Lunceford04}. This procedure ensures that the \textit{weighted} RCT distribution is similar to the observational one. For example, an RCT sample with low probability is weighted higher in order to offset the low chance of its inclusion into the trial. Investigators then estimate $f(\bm{X})$ using the derived weights. These methods unfortunately fail to accommodate \textit{strict} exclusion criteria which exclude specific patients with absolute certainty so that $1/\mathbb{P}(\bm{S}=1|T,\bm{X})=\infty$ in these cases. IPWEs therefore assume $\mathcal{S}_O \subseteq \mathcal{S}_R$, even though this rarely holds in practice. Weighting methods also use the large sample sizes of the observational data to estimate $\mathbb{P}(\bm{S}=1|T,\bm{X})$, but they ultimately rely on the underpowered RCT to estimate $f(\bm{X})$. Both of these limitations result in poor performance in practice.

A third line of algorithms improve CATE estimation by increasing the sample efficiency of RCTs through a variety of techniques. The methods for example exploit parameter sharing \citep{Ilse21}, feature selection \citep{Triantafillou20,Triantafillou21}, shared treatment effects \citep{Rosenman20} and calibration \citep{Zeng21}. All of these algorithms however require $\mathcal{S}_O \subseteq \mathcal{S}_R$ like IPWEs. As a result, they improve interpolation but continue to extrapolate poorly to the excluded patients.

A final set of methods specifically attempt to extrapolate by assuming certain functional forms linking the observational and RCT distributions. The earliest works assume a linear relation:
\begin{equation} \label{eq:OLT}
    f(\bm{X}) = g(\bm{X})\alpha + \delta,
\end{equation}
\noindent where we first learn the confounded $g(\bm{X}) \triangleq  \mathbb{E}(Y(1)|\bm{X},T=1) - \mathbb{E}(Y(0)|\bm{X},T=0)$ using the observational data. We then learn $f(\bm{X})$ by fitting the parameters $\alpha, \delta \in \mathbb{R}$ using linear regression on $\mathcal{S}_R \cap \mathcal{S}_O = \mathcal{S}_R$ \citep{Jackson17}. Another paper proposed the equation:
\begin{equation} \label{eq:2Step}
    f(\bm{X}) = g(\bm{X}) + \bm{X}\theta + \phi,
\end{equation}
where we again first learn $g(\bm{X})$ but then correct it using a linear term dependent on $\bm{X}$ \citep{Kallus18}. The linear correction may however perform poorly when $f(\bm{X})$ and $g(\bm{X})$ differ according to a non-linear function of $\bm{X}$.

In this paper, we take an alternative approach that combines the ideas of DD and the linear correction. A conditional version of DD uses the observational data to estimate the conditional average treatment effect on the treated (CATT), mathematically written as $\mathbb{E}(Y(1)-Y(0)|\bm{X},T=1)$, with four potentially non-linear functions of $\bm{X}$. The algorithm however relies on a parallel slopes assumption that may not apply in practice (details below). SDD relaxes the assumption by optimally re-weighting the components using additional RCT data in order to recover $f(\bm{X})$. SDD therefore improves upon DD by accommodating non-parallel slopes via a linear adjustment.

\section{Difference in Differences} \label{sec_DD}

\subsection{Unconditional Version}

Difference in Differences (DD) is a statistical technique for estimating the average treatment effect on the treated (ATT) $\mathbb{E}(Y(1)-Y(0)|T=1)$ using observational data \citep{Card93}. We consider two time steps: \textit{pre- and post-treatment administration} denoted by a binary variable $M$ equal to $0$ or $1$, respectively. We use the notation $Y_M(T)$ to mean the potential outcome of treatment assignment $T$ at time point $M$. Let $\bm{X} = \emptyset$ in this subsection.

\begin{figure*}
    \centering
    \begin{subfigure}[b]{0.45\textwidth}
    \centering
    \includegraphics[scale=0.11]{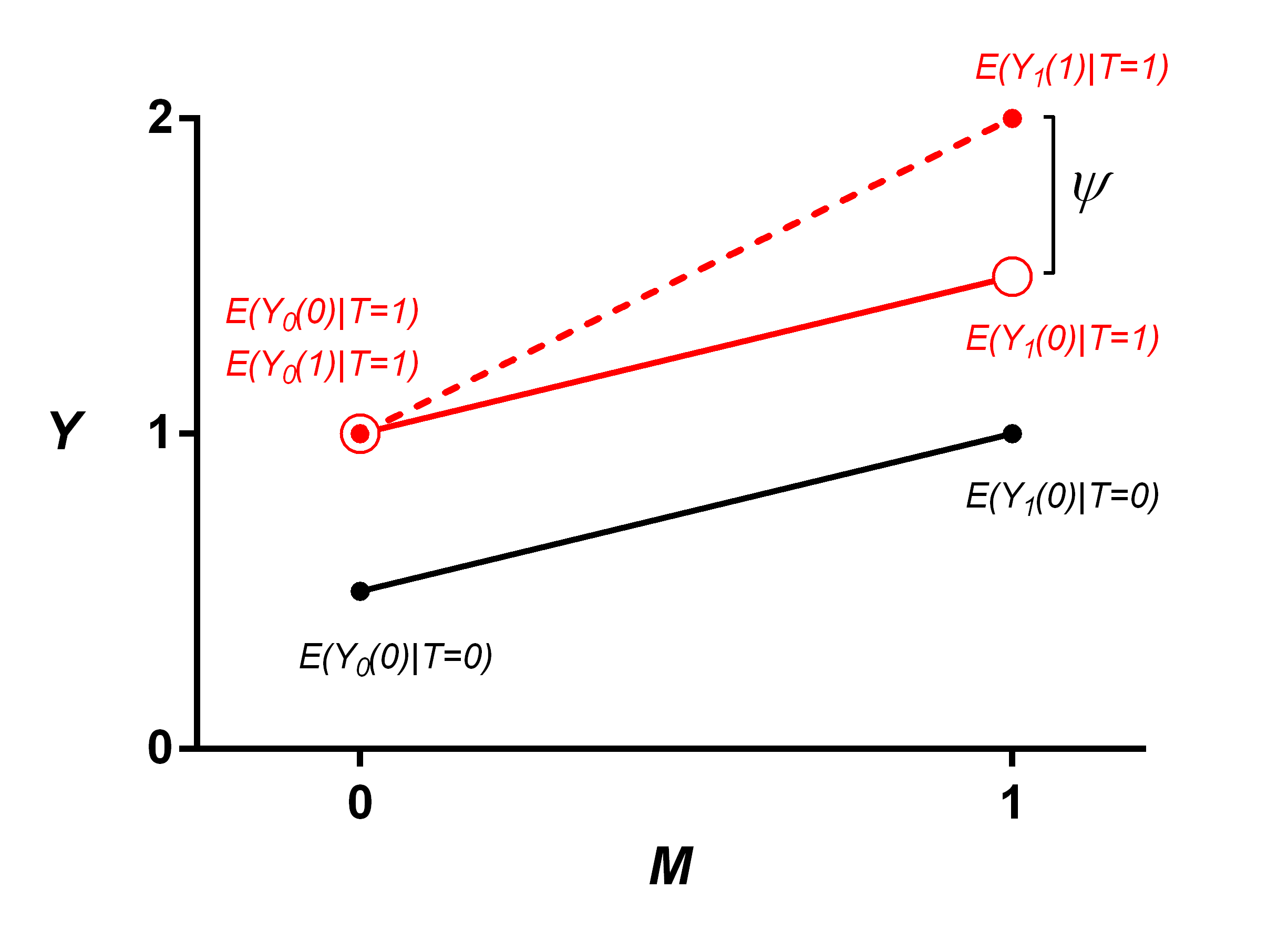}
    \caption{} \label{fig:DD}
    \end{subfigure}
    \begin{subfigure}[b]{0.45\textwidth}
    \centering
    \includegraphics[scale=0.69]{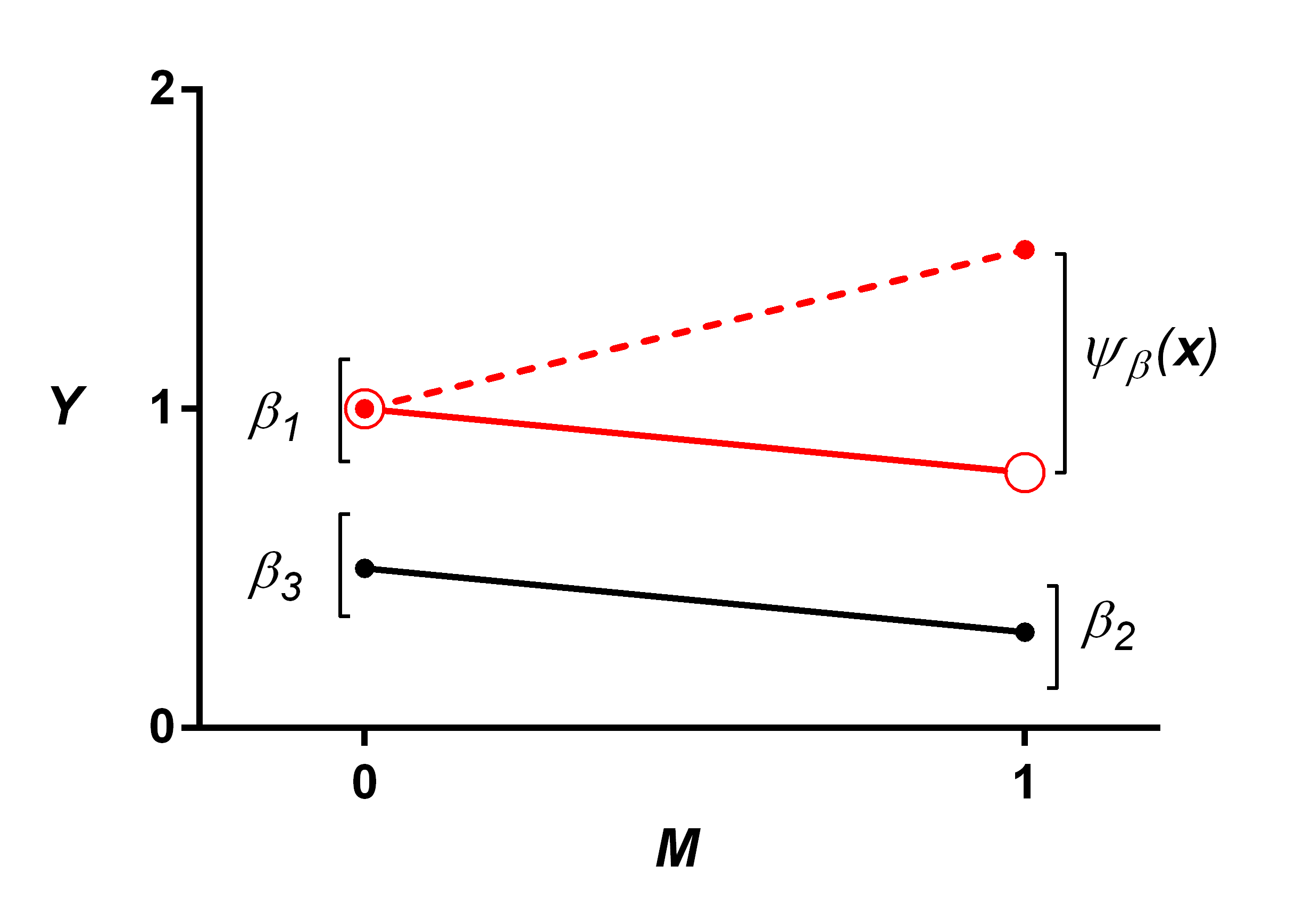}
    \caption{}  \label{fig:coef}
    \end{subfigure}
    \caption{Visualization of DD and SDD in the observational distribution. Red lines denote $T=1$ and black $T=0$. Filled circles are observable values, whereas unfilled circles are not. (a) Assumption \ref{as_preT} corresponds to the equality on the left side and Assumption \ref{as_slope} to the parallel slopes for the solid lines. (b) The coefficients move the corresponding dots up or down to allow the non-parallel slopes of Assumption \ref{as_NPslope}. This graph corresponds to a particular choice $\bm{X}=\bm{x}$ and may differ between patients.}
\end{figure*}

DD requires two main assumptions. The first states that the treatment does not have any effect before treatment administration:
\begin{assumptions1} \label{as_preT}
(No pre-treatment effect) We have $\mathbb{E}(Y_0(1)|\bm{X},T=1) = \mathbb{E}(Y_0(0)|\bm{X},T=1)$.
\end{assumptions1}
\noindent The second requires the following equality between treatment groups:
\begin{assumptions1} \label{as_slope}
(Parallel slopes) We have $\mathbb{E}(Y_1(0)|\bm{X},T=1) - \mathbb{E}(Y_0(0)|\bm{X},T=1) =  \mathbb{E}(Y_1(0)|\bm{X},T=0) - \mathbb{E}(Y_0(0)|\bm{X},T=0)$.
\end{assumptions1}
\noindent We can visualize these two assumptions graphically in Figure \ref{fig:DD} with $\bm{X}=\emptyset$ for the unconditional case. The first assumption means that the dotted and solid red lines meet at time point $M=0$. The second assumption ensures that the solid black and solid red lines are parallel. Notice that the parallel slopes only apply to the potential outcomes $Y_0(0)$ and $Y_1(0)$. Moreover, Assumption \ref{as_slope} implies a constant confounding effect across time because rearranging the equation reveals that the confounding effect $\mathbb{E}(Y_0(0)|\bm{X},T=1) - \mathbb{E}(Y_0(0)|\bm{X},T=0)$ at time point $M=0$ equals the confounding effect $\mathbb{E}(Y_1(0)|\bm{X},T=1) - \mathbb{E}(Y_1(0)|\bm{X},T=0)$ at $M=1$.

DD unfortunately cannot estimate the ATT $\mathbb{E}(Y_1(1)|T=1) - \mathbb{E}(Y_1(0)|T=1)$ directly because the potential outcome $Y_1(0)$ does not correspond to the assigned treatment $T=1$ in the second conditional expectation. We therefore cannot observe the samples needed to estimate  $\mathbb{E}(Y_1(0)|T=1)$ as represented by the unfilled circle on the right in Figure \ref{fig:DD}. DD instead computes a ``difference of differences'' using the observational data:
\begin{equation}
\begin{aligned}
    \psi = &\hspace{1mm}[\mathbb{E}(Y_1(1)|T=1) - \mathbb{E}(Y_0(1)|T=1)]\\ &- [\mathbb{E}(Y_1(0)|T=0) - \mathbb{E}(Y_0(0)|T=0)],
\end{aligned}
\end{equation}
where the potential outcomes always coincide with the assigned treatments. This estimand is labeled in Figure \ref{fig:DD} and equals the ATT under the aforementioned assumptions:
\begin{proposition1} \label{prop:ATT}
$\psi$ is equal to the ATT under Assumptions \ref{as_preT} and \ref{as_slope}.
\end{proposition1}
\begin{proof}
We prove the more general conditional case. We can write:
\begin{equation} \nonumber
\begin{aligned}
        &\mathbb{E}(Y_1(1) - Y_1(0)|\bm{X},T=1)\\
         \stackrel{\ref{as_preT}}{=} &\hspace{1mm}\mathbb{E}(Y_1(1)|\bm{X},T=1) - \mathbb{E}(Y_1(0)|\bm{X},T=1)\\ &+ \mathbb{E}(Y_0(0)|\bm{X},T=1) - \mathbb{E}(Y_0(1)|\bm{X},T=1)\\
         = &\hspace{1mm}\mathbb{E}(Y_1(1)|\bm{X},T=1) - [\mathbb{E}(Y_1(0)|\bm{X},T=1)\\ &- \mathbb{E}(Y_0(0)|\bm{X},T=1)] - \mathbb{E}(Y_0(1)|\bm{X},T=1)\\
        \stackrel{\ref{as_slope}}{=} &\hspace{1mm}\mathbb{E}(Y_1(1)|\bm{X},T=1) - [\mathbb{E}(Y_1(0)|\bm{X},T=0)\\ &- \mathbb{E}(Y_0(0)|\bm{X},T=0)] - \mathbb{E}(Y_0(1)|\bm{X},T=1)\\
        = &\hspace{1mm}[\mathbb{E}(Y_1(1)|\bm{X},T=1) - \mathbb{E}(Y_0(1)|\bm{X},T=1)]\\ &- [\mathbb{E}(Y_1(0)|\bm{X},T=0) - \mathbb{E}(Y_0(0)|\bm{X},T=0)]\\
        = &\hspace{1mm}\psi,
\end{aligned}
\end{equation}
where the numbers above the equality signs correspond to the respective assumptions. The conclusion follows by setting $\bm{X} = \emptyset$.
\end{proof}
\noindent Assumptions \ref{as_preT} and \ref{as_slope} thus allow us to decompose the ATT into a combination of four conditional expectations whose treatment assignment $T$ and potential outcome $Y_M(T)$ always match.

\subsection{Conditional Version} \label{sec_CDD}

We can extend DD to the conditional setting in order to estimate the conditional average treatment effect on the treated (CATT) $\mathbb{E}(Y_1(1)-Y_1(0)|\bm{X},T=1)$ \citep{Abadie05}. We write the Conditional Difference in Differences (CDD) formula as follows using the pre-treatment covariates $\bm{X} \not = \emptyset$:
\begin{equation} \nonumber
\begin{aligned}
    \psi(\bm{X}) = &\hspace{1mm}[\mathbb{E}(Y_1(1)|\bm{X},T=1) - \mathbb{E}(Y_0(1)|\bm{X},T=1)]\\ &- [\mathbb{E}(Y_1(0)|\bm{X},T=0) - \mathbb{E}(Y_0(0)|\bm{X},T=0)].
\end{aligned}
\end{equation}
The function $\psi(\bm{X})$ corresponds to the CATT using the same argument as in the previous section:
\begin{proposition1}
$\psi(\bm{X})$ is equal to the CATT under Assumptions \ref{as_preT} and \ref{as_slope}.
\end{proposition1}
\noindent CDD thus imposes the parallel slopes assumption like DD.

\section{Synthesized Difference in Differences} \label{sec_SDD}

\subsection{Non-Parallel Slopes}
The parallel slopes assumption may unfortunately fail to hold in practice when the confounding effect shifts across time, such as in the example of Section \ref{sec_intro}. We therefore wish to accommodate non-parallel slopes in order to increase the flexibility of CDD. We relax Assumption \ref{as_slope} as follows:
\begin{assumptions1} \label{as_NPslope}
(Non-parallel slopes) There exists coefficients $\beta \triangleq (\beta_1, \beta_2,\beta_3) \geq 0$ such that $\mathbb{E}(Y_1(0)|\bm{X},T=1) - \mathbb{E}(Y_0(0)|\bm{X},T=1)\beta_1 =  \mathbb{E}(Y_1(0)|\bm{X},T=0)\beta_2 - \mathbb{E}(Y_0(0)|\bm{X},T=0)\beta_3$.
\end{assumptions1}
\noindent Assumption \ref{as_slope} requires that $\beta = (1,1,1)$, but Assumption \ref{as_NPslope} allows any positive $\beta$ in order to model non-parallel slopes as shown in Figure \ref{fig:coef}. The coefficients allow us to move the corresponding dots up or down by magnitudes $\beta$ for all patients. Note that we restrict the coefficients to positive numbers in order to preserve the equality of \textit{differences} requirement in Assumption \ref{as_slope}; in other words, we prevent any differences from suddenly switching into sums by negative multiplication and vice versa. We also do not add a (strictly) positive weight to the first term on the left in Assumption \ref{as_NPslope} because we can appropriately normalize the equation by division so that the weight is equal to one. Assumption \ref{as_NPslope} ultimately allows us to claim:
\begin{proposition1} \label{thm_psiCATT}
The CATT is equal to: 
\begin{equation} \nonumber
\begin{aligned}
    \psi_\beta(\bm{X}) \triangleq &\hspace{1mm}[\mathbb{E}(Y_1(1)|\bm{X},T=1) - \mathbb{E}(Y_0(1)|\bm{X},T=1)\beta_1]\\ &- [\mathbb{E}(Y_1(0)|\bm{X},T=0)\beta_2 - \mathbb{E}(Y_0(0)|\bm{X},T=0)\beta_3],
\end{aligned}
\end{equation}
under Assumptions \ref{as_preT} and \ref{as_NPslope}.
\end{proposition1}

Investigators are nevertheless often more interested in estimating the CATE than the CATT because the CATT $\mathbb{E}(Y_1(1)-Y_1(0)|\bm{X},\\T=1)$ only applies to patients assigned to $T=1$, whereas the CATE $\mathbb{E}(Y_1(1)-Y_1(0)|\bm{X})$ applies to everyone regardless of treatment assignment. Providing additional flexibility with $\beta$ also raises the question: how do we choose $\beta$ among all the possibilities? Both of these issues inspire the following assumption:
\begin{assumptions1} \label{as_CATEslope}
(Linear adjustment) There exist constants $\beta \geq 0$ such that $\psi_\beta(\bm{X}) = f(\bm{X})$.
\end{assumptions1}
\noindent In other words, we assume that the CATE is reachable by a positively weighted combination of the component conditional expectations of the CATT. 

The above assumption differs from those used in the previous works summarized in Section \ref{sec_RW}. Equation \eqref{eq:2Step} assumes that the covariates are properly chosen a priori so that the linear relation holds, while Equation \eqref{eq:OLT} preserves the shape of $g(\bm{X})$. Assumption \ref{as_CATEslope} can analogously preserve the shape of the CATT. However, it more specifically considers the CATT to be a reasonable approximation of the CATE, so long as we can accommodate non-parallel slopes by optimizing the coefficients $\beta$ as detailed in the next subsection.

\begin{algorithm}[b]
 \nonl \textbf{Input:} RCT data, observational data, test points $\mathcal{T}$\\
 \nonl \textbf{Output:} $\widehat{\psi}_\beta(\bm{X})$ on $\mathcal{T}$\\
 \BlankLine

Estimate $f(\bm{X})$ using the RCT data \label{alg:iT}\\
Estimate $\mathbb{E}(Y_1(1)|\bm{X},T=1)$ and each entry of $H(\bm{X})$ using the observational data \label{alg:G}\\
Linearly regress $\widehat{k}(\bm{X})$ on $\widehat{H}(\bm{X})$ using the RCT data and then predict $\widehat{\psi}_\beta(\bm{X})$ on $\mathcal{T}$ \label{alg:eT}

 \caption{Synthesized Difference in Differences} \label{alg_SDD}
\end{algorithm}

\subsection{The Algorithm}

We now propose the Synthesized Difference in Differences (SDD) algorithm to exploit Assumption \ref{as_CATEslope} while simultaneously generalizing the CATE to the entirety of $\mathcal{S}_O$. We provide a summary in Algorithm 1.

SDD first learns the CATE on $\mathcal{S}_R$ using the RCT data in Line \ref{alg:iT} by estimating:
\begin{equation} \label{eq:RCTreg} 
\mathbb{E}(Y_1(1) | \bm{X},T=1,\bm{S}=1) - \mathbb{E}(Y_1(0) | \bm{X},T=0,\bm{S}=1),
\end{equation}
using two non-linear regressions. The algorithm first estimates $\mathbb{E}(Y_1(1)|\bm{X},T=1,\bm{S}=1)$ and then estimates $\mathbb{E}(Y_1(0)|\bm{X},T=0,\bm{S}=1)$ with the RCT; the difference of the two terms corresponds to the RCT estimate $\widehat{f}(\bm{X})$ by Assumptions \ref{ax_RCT} and \ref{as_SB}. 

We can also rewrite the CATE in an alternative form with the RCT distribution as a function of the row vector $[-\mathbb{E}(Y_0(1)|\bm{X},T=1), -\mathbb{E}(Y_1(0)|\bm{X},T=0), \mathbb{E}(Y_0(0)|\bm{X},T=0)] \triangleq H(\bm{X})$:
\begin{theorem1} \label{thm:RCT}
Expression \eqref{eq:RCTreg} is equivalent to $\mathbb{E}(Y_1(1)|\bm{X},T=1) + H(\bm{X})\beta$ in the RCT distribution under Assumptions \ref{ax_RCT}, \ref{as_SB} and \ref{as_CATEslope}.
\end{theorem1}
\noindent SDD uses the observational data to estimate the above form by first approximating $\mathbb{E}(Y_1(1)|\bm{X},T=1)$ and $H(\bm{X})$ in Line \ref{alg:G}. 

The algorithm then takes the four fitted conditional expectations and estimates $\beta$ by regressing $\widehat{k}(\bm{X})=\widehat{f}(\bm{X}) - \widehat{\mathbb{E}}(Y_1(1)|\bm{X},T=1)$ on $\widehat{H}(\bm{X})$ in Line \ref{alg:eT} using the RCT data:
\begin{equation} \label{eq:creg}
\begin{aligned}
    \min_{\beta} &\frac{1}{4m} \sum_{i=1}^{2m} (\widehat{k}(\bm{x}_i) - \widehat{H}(\bm{x}_i)\beta)^2 + \frac{\lambda}{4m} \| \beta -  1 \|_2^2\\
    &\hspace{15mm} s.t. \hspace{2mm} 0 \leq \beta \leq 3,
\end{aligned}
\end{equation}
where we assume $m$ samples per treatment group in the RCT for notational convenience. The hard constraint and regularization term are not necessary from a theoretical standpoint, but they improve finite sample performance by ensuring (1) positivity of the coefficients by Assumption \ref{as_CATEslope}, and (2) that the coefficients do not deviate too far from those of CDD. The constraint and regularization also prevent SDD from considering very poor estimates of the true CATE in difficult cases where most patients are excluded from the RCT. We will analyze the effects of Expression \eqref{eq:creg} in the next section as compared to unconstrained linear regression. SDD finally outputs $\widehat{\psi}_\beta(\bm{X}) = \widehat{\mathbb{E}}(Y_1(1)|\bm{X},T=1) + \widehat{H}(\bm{X})\widehat{\beta}$ in accordance with Theorem \ref{thm:RCT}.

In summary, SDD first estimates $f(\bm{X})$ using the RCT. The algorithm then uses the observational dataset to approximate the four non-linear conditional expectations $\mathbb{E}(Y_1(1)|\bm{X},T=1)$ and $H(\bm{X})$ in $\psi_{\beta}(\bm{X})$. Finally, SDD learns $\beta$ in $\psi_{\beta}(\bm{X})$ by regressing $\widehat{f}(\bm{X}) - \widehat{\mathbb{E}}(Y_1(1)|\bm{X},T=1)$ on $\widehat{H}(\bm{X})$. Combining $\widehat{\mathbb{E}}(Y_1(1)|\bm{X},T=1)$, $\widehat{H}(\bm{X})$ and $\widehat{\beta}$ yields $ \widehat{\psi}_{\beta}(\bm{X})$ such that  $\widehat{\psi}_{\beta}(\bm{X})\approx \widehat{f}(\bm{X})$.

\subsection{Consistency}
We can theoretically justify SDD in terms of its consistency. We first have:
\begin{theorem1} (Fisher consistency) \label{thm:fisher}
If $\mathbb{E}(H(\bm{X})^TH(\bm{X}))$ is invertible, then the population version of SDD recovers $f(\bm{X})$ for any $\bm{X} \in \mathcal{S}_O$ under Assumptions \ref{as_support}-\ref{as_SB} and \ref{as_CATEslope}.
\end{theorem1}
We next establish SDD's rate of convergence. Assume that the sample size $n$ is the same for each treatment in the observational data for convenience.
\begin{theorem1} (Convergence rate) \label{thm_conv}
Consider Assumptions \ref{as_support}-\ref{as_SB} and \ref{as_CATEslope}. Further assume:
\begin{enumerate}
\item The regression procedures in Lines \ref{alg:iT} and \ref{alg:G} converge to their estimands at rate $O_p(s(m)) = o_p(1)$ on $\mathcal{S}_R$ and $O_p(r(n)) = o_p(1)$ on $\mathcal{S}_O$, respectively.
\item $H(\bm{X})^Tk(\bm{X})$ has finite covariance.
\item $\mathbb{E}(H(\bm{X})^TH(\bm{X}))$ is finite and invertible. 
\end{enumerate}
Then SDD approximates the true CATE for any $\bm{X} \in \mathcal{S}_O$ at rate $O_p(s(m)) + O_p(r(n)) + O_p(1/\sqrt{m})$ corresponding to Lines \ref{alg:iT}, \ref{alg:G} and \ref{alg:eT}, respectively.
\end{theorem1}
\noindent SDD is thus limited by the slowest of the three regressions. The algorithm can also use many different regression procedures in practice. We use kernel ridge regression in our implementation to more specifically achieve the optimal rate $O_p(1/\sqrt{n}) + O_p(1/\sqrt{m})$ and obtain good empirical performance even in the small sample regime seen with RCTs \citep{Hable12}.

\section{Experiments} \label{sec_exp}

\subsection{Algorithms}
We compared the following six algorithms on synthetic and real data:
\begin{enumerate}
    \item SDD
    \item CDD \citep{Abadie05}
    \item 2Step \citep{Kallus18}
    \item Outer Linear Transform (OLT) \citep{Jackson17}
    \item Regression with observational data only (OBS) \citep{Rubin78,Rosenbaum83,Pearl09}
    \item Regression with RCT data only \citep{Hahn98,Heckman97,Heckman98}
\end{enumerate}
SDD, 2Step and OLT use observational and RCT data, whereas CDD only uses observational data. Regression using observational data only assumes unconfoundedness, whereas regression using RCT data only assumes no strict exclusion criteria; we will use the acronyms OBS and RCT to refer to these two methods, when it is clear that we mean the algorithms. Out of the box, each algorithm in the above list uses different regressors that can be swapped to any method of choice. We therefore replaced all non-linear regressions with kernel ridge regression using universal first degree INK-spline kernels in order to isolate the performance of each algorithm independent of the underlying regressor \citep{Izmailov13}. We chose this setup due to (1) the strong performance of the kernel method even with the limited sample sizes of RCTs, and (2) the optimal theoretical convergence rate according to Theorem \ref{thm_conv}. We selected the $\lambda$ hyperparameter from the set $\{$1E-8$,\dots,$1E-1$\}$ for kernel ridge regression and, with SDD, for Expression \eqref{eq:creg} using five-fold cross-validation.

\subsection{Synthetic Data}
\begin{table*}[t]
\begin{subtable}{0.45\textwidth}   
\centering
\begin{tabular}{ccccccc}
\hhline{=======}
\multicolumn{1}{l}{} & 0.00        & 0.25        & 0.50 & 0.75 & 0.90 & 0.95 \\ \hline
SDD                                           & \textbf{0.0753}                     & \textbf{0.0780}                     & \textbf{0.0967}              & \textbf{0.1151}              & \textbf{0.1402}              & \textbf{0.1746}              \\
CDD                                           & 0.2141                              & 0.2064                              & 0.2250                       & 0.2220                       & 0.2102                       & 0.2195                       \\
2Step                                         & 0.4214                              & 0.4291                              & 0.5001                       & 0.6214                       & 0.6635                       & 0.7261                       \\
OLT                                           & 0.4360                              & 0.4390                              & 0.4987                       & 0.5508                       & 0.5842                       & 0.7633                       \\
OBS                                           & 0.3746                              & 0.3693                              & 0.3714                       & 0.3617                       & 0.3678                       & 0.3784                       \\
RCT                                           & 0.1063                              & 0.1550                              & 0.3106                       & 0.5237                       & 0.8314                       & 0.9764                       \\ \hdashline
-PRE                                          & \multicolumn{1}{l}{0.3373}          & \multicolumn{1}{l}{0.3412}          & \multicolumn{1}{l}{0.3726}   & \multicolumn{1}{l}{0.4059}   & \multicolumn{1}{l}{0.3957}   & \multicolumn{1}{l}{0.4449}   \\
-CON                                          & \multicolumn{1}{l}{\textbf{0.0749}} & \multicolumn{1}{l}{\textbf{0.0849}} & \multicolumn{1}{l}{0.1249}   & \multicolumn{1}{l}{0.1582}   & \multicolumn{1}{l}{0.2004}   & \multicolumn{1}{l}{0.2399}  \\ 
\hhline{=======}
\end{tabular}
\caption{} \label{table:prop}
\end{subtable}
\hspace{15mm}\begin{subtable}{0.45\textwidth}   
\centering
\begin{tabular}{ccccc}
\hhline{=====}
\multicolumn{1}{l}{} & 1  & 3  & 6  & 10          \\ \hline
SDD                                           & \textbf{0.0242}            & \textbf{0.0588}                     & \textbf{0.1538}                     & \textbf{0.3007}                     \\
CDD                                           & 0.0602                     & 0.1460                              & 0.3088                              & 0.5418                              \\
2Step                                         & 0.1945                     & 0.4249                              & 0.7421                              & 1.1587                              \\
OLT                                           & 0.2638                     & 0.3752                              & 0.6681                              & 1.0543                              \\
OBS                                           & 0.1799                     & 0.4429                              & 0.8258                              & 1.2931                              \\
RCT                                           & 0.1813                     & 0.2276                              & 0.3831                              & 0.6647                              \\ \hdashline
-PRE                                          & \multicolumn{1}{l}{0.1018} & \multicolumn{1}{l}{0.2837}          & \multicolumn{1}{l}{0.5018}          & \multicolumn{1}{l}{0.7289}          \\
-CON                                          & \multicolumn{1}{l}{0.0772} & \multicolumn{1}{l}{\textbf{0.0618}} & \multicolumn{1}{l}{\textbf{0.1579}} & \multicolumn{1}{l}{\textbf{0.3075}}\\
\hhline{=====}
\end{tabular}
\caption{} \label{table:dim}
\end{subtable}
\caption{Median values of the MSEs. Columns correspond to (a) the proportion of excluded patients, and (b) the number of dimensions. Bolded values correspond to the best performance within each column.}
\end{table*}

We first generated the RCT data with the following model:
\begin{equation} \nonumber
\begin{aligned}
    Y_1(1) &= \mathbb{E}(Y_1(1)|\bm{X},T=1) - \mathbb{E}(Y_0(1)|\bm{X},T=1)\beta_1 + \varepsilon_1,\\
    Y_1(0) &= \mathbb{E}(Y_1(0)|\bm{X},T=0)\beta_2 - \mathbb{E}(Y_0(0)|\bm{X},T=0)\beta_3 + \varepsilon_0,
\end{aligned}
\end{equation}
using independent error terms $\varepsilon \sim \sqrt{0.1}\mathcal{N}(0,I)$. We also generated the observational data with:
\begin{equation} \nonumber
    Y_M(T) = \mathbb{E}(Y_M(T)|\bm{X},T)+ \varepsilon_{MT}.
\end{equation}
The CATE corresponds to $\psi_{\beta}(\bm{X})$ according to Assumption \ref{as_CATEslope} with each entry in $\beta$ sampled independently from $\mathcal{U}([0.5,1.5])$ in order to model shifting confounding effects violating the parallel slopes assumption. Let $Z = \sum_i X_i$. We sampled the four conditional expectations in $\psi_{\beta}(\bm{X})$ independently and uniformly from the set of post-nonlinear functions $\Big\{Z, Z \Phi(Z), \sum_i \textnormal{exp}(-Z^2), \textnormal{tanh}(Z)\Big\}$ for each value of $M$ and each value of $T$. We simulated each entry of $\bm{X}$ from mutually independent $\mathcal{U}(-1,1)$. We generated $1000$ samples for the observational data split evenly between treatments and time steps. We also generated $100$ samples for the RCT data split evenly between treatments. We repeated the above procedure 500 times for 1, 3, 6 and 10 variables in $\bm{X}$. We imposed strict exclusion criteria in the RCT ranging from mild to severe by excluding $r = 0, 25, 50, 75, 90$ or $95\%$ patients with $X_1 \sim \mathcal{U}(-1+0.02r,1)$. We therefore generated a total of $500 \times 4 \times 6 = 12000$ independent datasets. We compared the algorithms by assessing their median mean squared errors (MSEs) to the true CATE $f(\bm{X})$ on all observational samples with Mood's median test; we use the median rather than the mean due to the positive skewness of the recovered histograms.

\subsubsection{Performance}

\begin{figure*}
\begin{subfigure}[]{0.45\textwidth}
    \centering
    \includegraphics[scale=0.65]{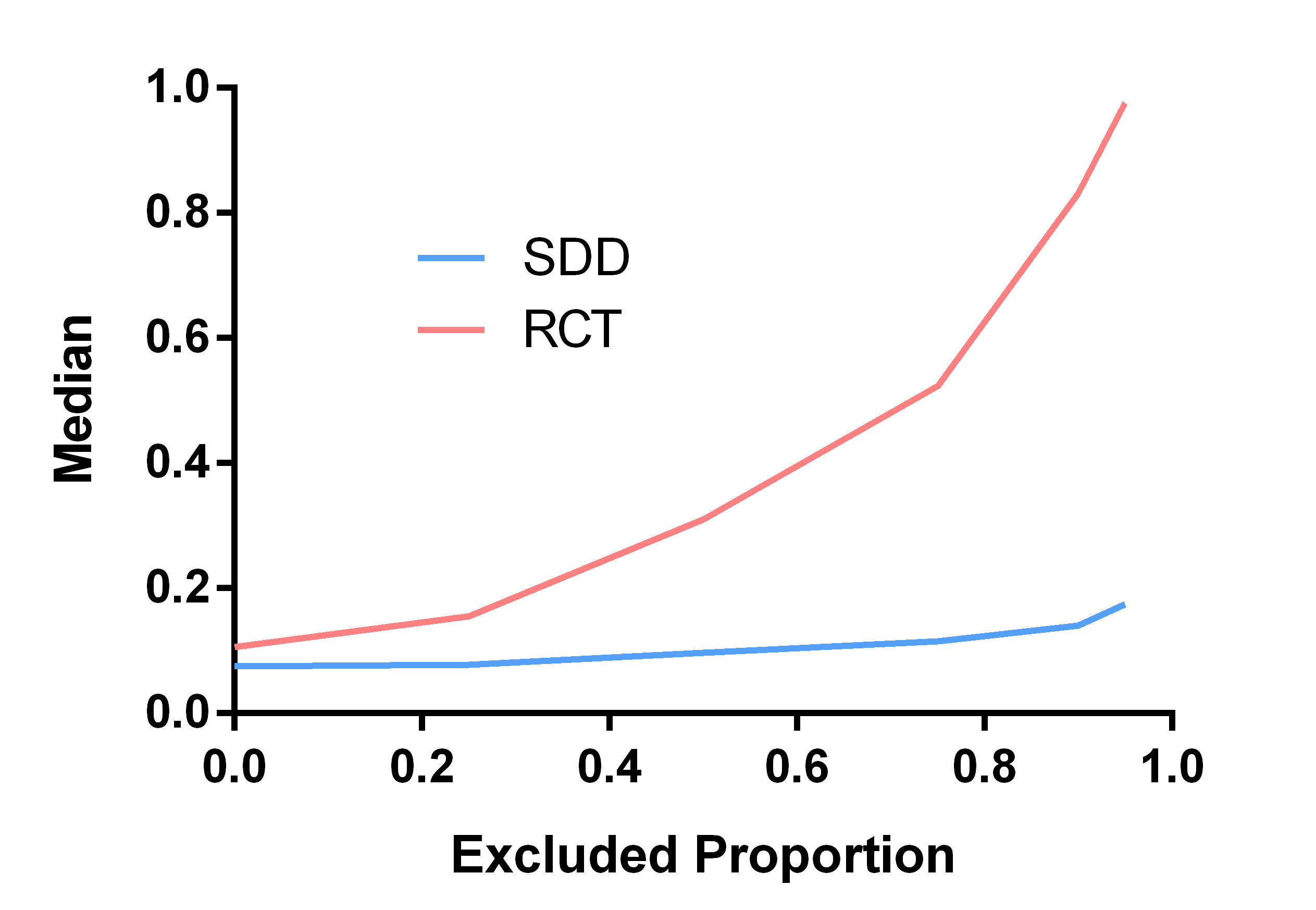}
    \caption{} \label{fig:SDDvRCT}
\end{subfigure}
\begin{subfigure}[]{0.45\textwidth}
    \centering
    \includegraphics[scale=0.65]{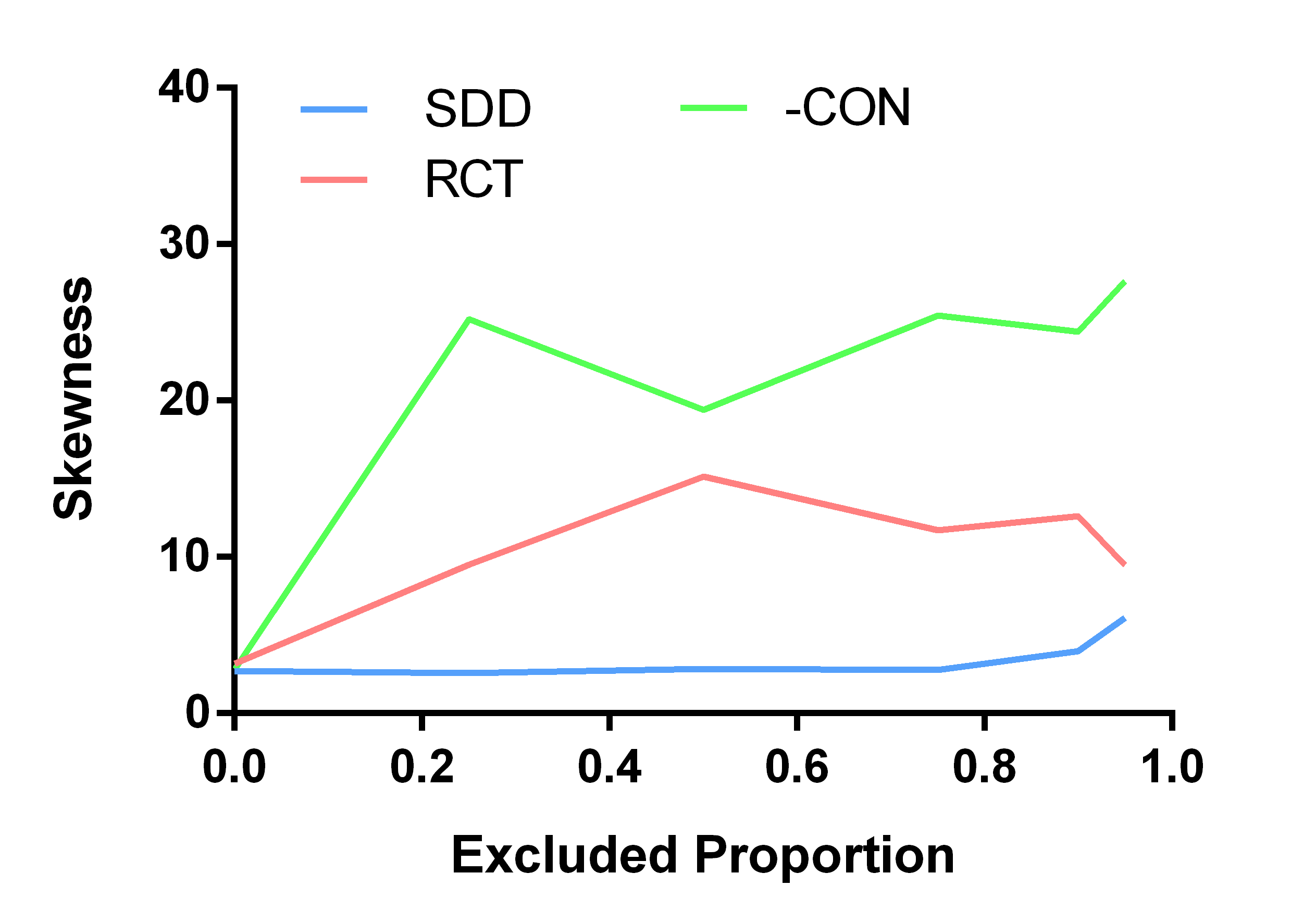}
    \caption{} \label{fig:SDDvRCTvCON}
\end{subfigure}
\caption{Visualization of (a) the median MSE values of SDD vs. RCT, and (b) the skewness of the MSE values of SDD vs. RCT and -CON. Estimating the CATE from the RCT alone becomes unreliable with more stringent exclusion criteria in (a). Constrained regression improves the estimated model by preventing catastrophic failures, or preventing skewness to the right, in (b).}
\end{figure*}

We summarize the median MSEs in Table 1. Bolded numbers correspond to the best performing algorithms that reached significance at a Bonferonni corrected threshold of 0.05/5. Table 1 (a) reports the results across proportion of excluded patients from the RCT. The left hand column corresponds to $0\%$ excluded patients while the right to $95\%$ excluded; the problem therefore becomes more difficult as a higher proportion of the patient population cannot enter the RCT. SDD outperformed all five of its competitors across all proportions. The RCT only algorithm came in second place for $0$ and $25\%$ excluded. RCT however failed to take advantage of the larger sample sizes in the observational data, and its performance gradually deteriorated as a higher proportion of patients could not enter the clinical trial (Figure \ref{fig:SDDvRCT}). Sometimes RCT generated very inaccurate estimates of the true CATE as seen in Figure \ref{fig:SDDvRCTvCON}, where the RCT histogram of MSE values had a higher positive skew towards larger MSE values than the SDD histogram. This is expected behavior because most non-parametric machine learning methods interpolate but cannot extrapolate. We finally assessed the results across dimensions of $\bm{X}$ (Table 1 (b)); SDD outperformed all other algorithms in this case as well.

\subsubsection{Ablation Studies}

We performed two ablation studies to assess the components of SDD. We first eliminated the pre-treatment administration time step (-PRE) to see if SDD can perform well with only post-treatment administration RCT and observational data. This modification unfortunately caused a deterioration in performance across all excluded proportions and dimensions (Tables 1 (a) and 1 (b)). We conclude that pre-treatment administration data improves performance.

We next replaced the constrained regression procedure shown in Expression \eqref{eq:creg} with ordinary least squares (-CON). -CON performed poorly with large proportions of excluded patients (Table 1 (a)) because the algorithm often recovered very inaccurate estimates of the MSE in these challenging settings. We visualize this phenomenon in Figure \ref{fig:SDDvRCTvCON}, where the -CON histogram of MSE values again has increasing positive skew with more excluded patients. SDD on the other hand controls the skew even with increasing proportions. We conclude that the constrained regression is critical for preventing catastrophic failures in difficult cases.

\subsection{Real Data}
Evaluating the algorithms on real data is difficult because we rarely have access to the ground truth CATE across the entire clinical population. Fortunately, scientists have performed a handful of large, expensive, multi-institutional RCTs with few exclusion criteria. We take advantage of these RCTs by splitting the sample on common exclusion criteria in order to mimic a more common exclusionary RCT. We then generate observational data by enforcing common prescribing patterns using the pre-treatment administration data.

\begin{figure*}
\begin{subfigure}[]{0.45\textwidth}
    \centering
    \includegraphics[scale=0.65]{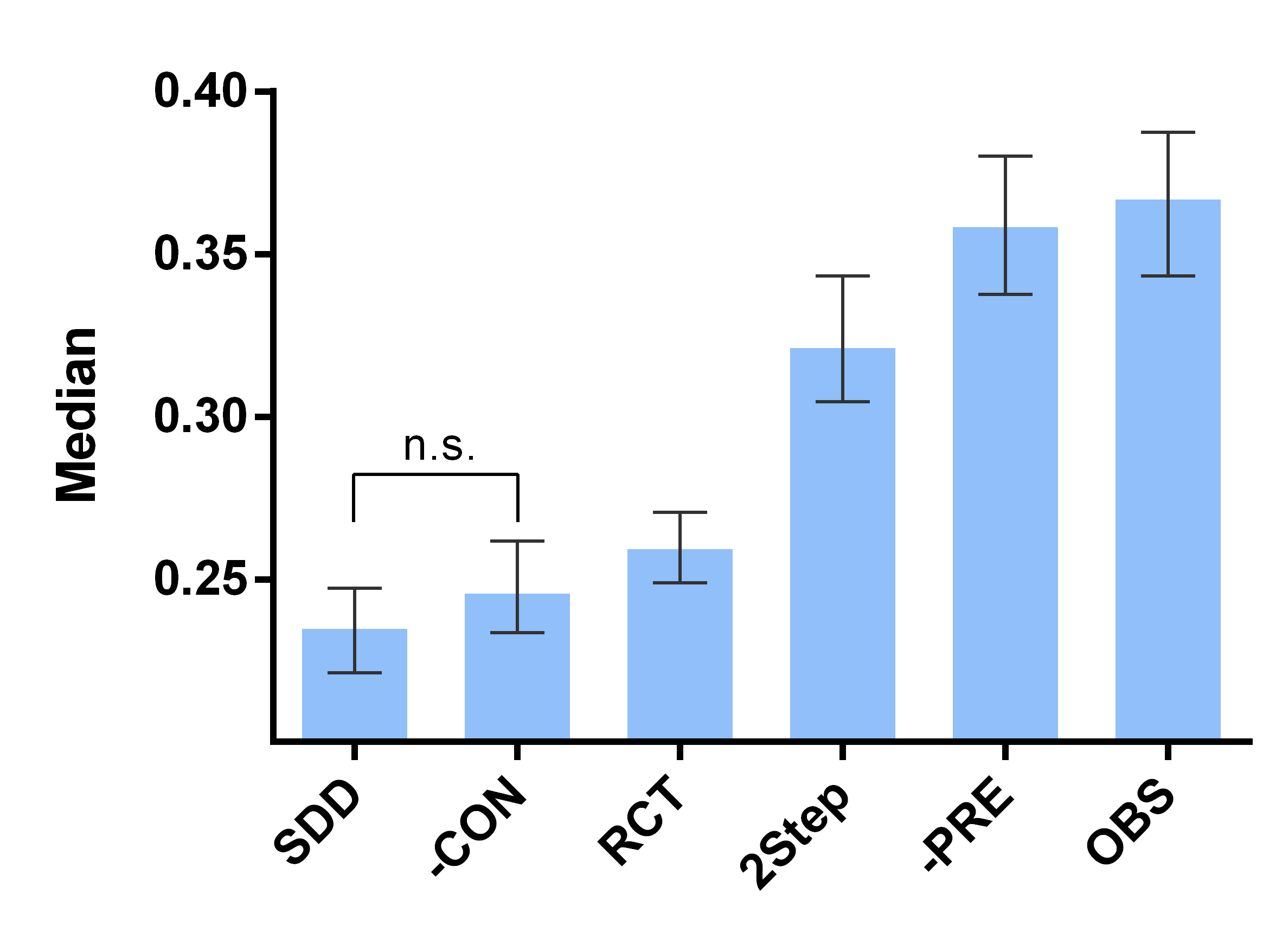}
    \caption{} \label{fig:STARD}
\end{subfigure}
\begin{subfigure}[]{0.45\textwidth}
    \centering
    \includegraphics[scale=0.65]{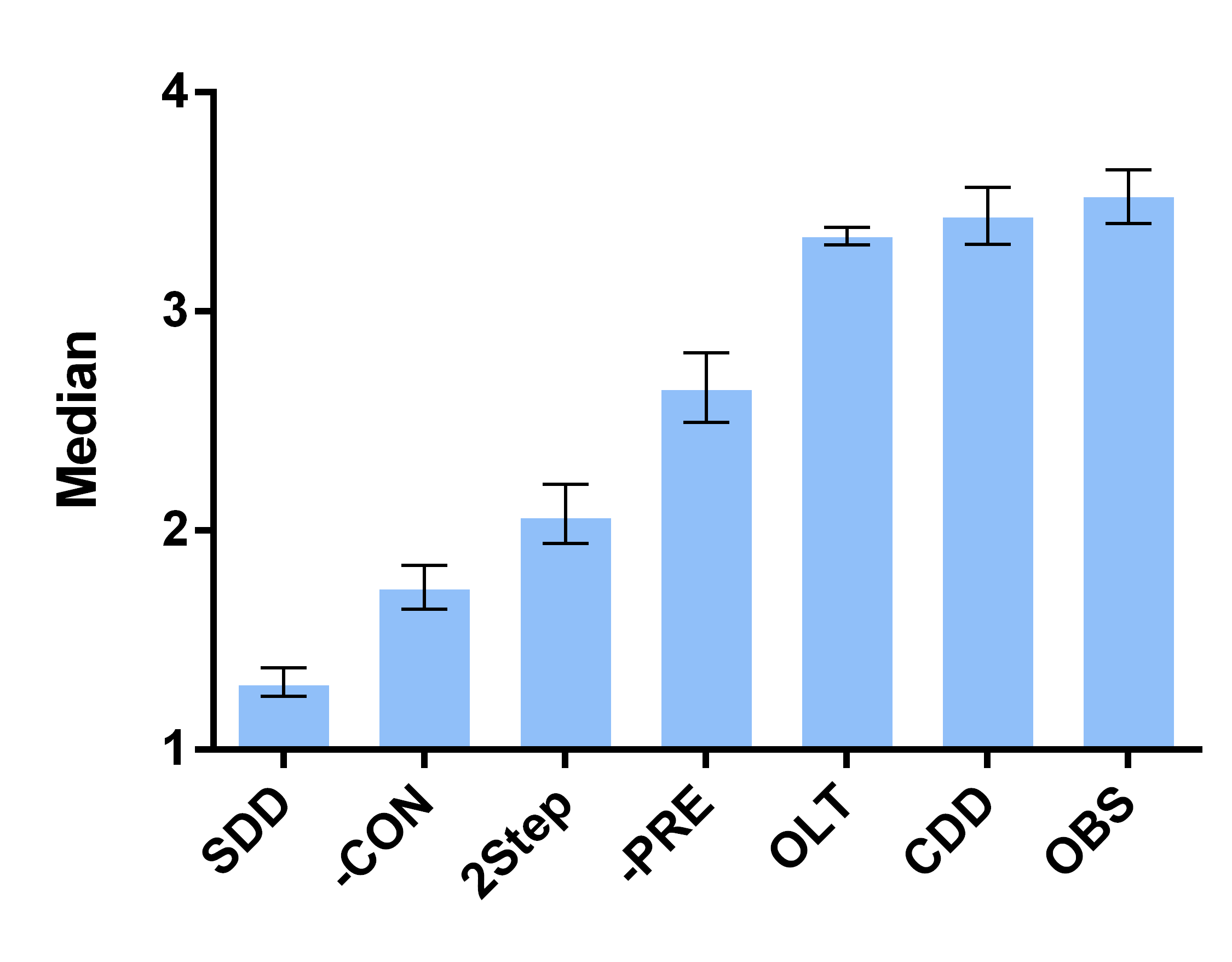}
    \caption{} \label{fig:TEOSS}
\end{subfigure}
\caption{Median MSE values with (a) STAR*D and (b) CATIE and TEOSS. Error bars correspond to 95\% confidence intervals. We exclude algorithms with very high median MSEs.}
\end{figure*}

\subsubsection{STAR*D}
The Sequenced Treatment Alternatives to Relieve Depression (STAR*D) was a large multi-million dollar RCT performed to assess the effect of anti-depressants and cognitive behavioral therapy on major depressive disorder \citep{Trivedi06,Rush06,Warden07}. The investigators assessed treatment response using the QIDS-SR score, a standardized self-reported measure of depressive symptoms. STAR*D ultimately included four levels of sequential treatment assignment. 

We used the second level of STAR*D because the level had a large sample size and tested the effects of buproprion ($T=1$) vs. venlafaxine and sertraline ($T=0$). Buproprion is known to have a unique effect on hypersomnia. We therefore set the week 6 hypersomnia sub-score of QIDS-SR as the outcome variable to give sufficient time for the drugs to elicit differential effects while minimizing patient dropout. We used the other sleep variables in the QIDS-SR at baseline as predictors $\bm{X}$, including sleep onset insomnia, mid-noctunal insomnia and early morning insomnia. We finally treated this entire STAR*D dataset as the population from which we estimated the ground truth CATE. 

We generated observational data by imposing confounding on the STAR*D dataset. We  kept all samples with $T=0$ but eliminated samples with a baseline QIDS-SR hypersomnia score less than one for $T=1$. This mimics real-world prescribing patterns, where physicians prefer to give buproprion to patients suffering from hypersomnia. This excluded 19.3\% of patients, reducing the STAR*D sample size from 388 to 313.

We generated RCT data by imposing exclusion criteria on the STAR*D dataset. We first identified current psychotropic use and substance use as the two most common exclusion criteria in clinical trials of major depression not already implemented in STAR*D \citep{Blanco17}. We therefore excluded individuals meeting at least one of those two criteria. This excluded 38.9\% of patients, reducing the STAR*D sample size from 388 to 237.

In summary, we introduced confounding using hypersomnia scores and selection bias using current psychotropic and substance use. We then ran all of the algorithms on $2000$ bootstraps of the derived observational and RCT datasets. We report the results in Figure \ref{fig:STARD}, where we have ordered the algorithms from lowest to highest median MSE from the ground truth CATE. We do not report CDD and OLT in the figure because they performed poorly (median MSEs 0.729 and 0.610 respectively). SDD and -CON attained the lowest median MSEs. RCT came in third place despite excluding 38.9\% of patients. We therefore conclude that the non-linear regressor extrapolated reasonably well in this setting. We will however see that RCT can fail catastrophically in the next example due to unsupervised extrapolation.

\subsubsection{CATIE and TEOSS}
 Many clinical trials exclude children. We therefore evaluated the algorithms on two clinical trials investigating the effects of anti-psychotics on schizophrenia-spectrum disorders in two different age groups - adults and children. CATIE recruited adults at least 18 years old, while TEOSS recruited children up to 19 years old \citep{Mcevoy05,Stroup09,Sikich08}. We used the 530 patients in CATIE as the RCT data and then tested on the 62 patients in TEOSS to assess non-overlapping distributions.
 
Clinicians prefer olanzapine over risperidone in agitated patients because olanzapine is more sedating. We therefore used the excitement subscore of the standardized Positive and Negative Syndrome Scale (PANSS) at week 4 as the outcome. We predicted differences in the outcome between olanzapine ($T=1$) versus risperidone ($T=0$) using age and the hostility sub-score as predictors $\bm{X}$. 

We created the observational dataset by combining the CATIE and TEOSS trials. We then introduced confounding by excluding patients with an excitement subscore less than or equal to two for olanzapine, and greater than two for risperidone. This process eliminated 46.7\% of patients.
 
In summary, we introduced confounding using excitement scores and selection bias using age. We report the results of $2000$ bootstraps in Figure \ref{fig:TEOSS}. SDD again achieved the lowest median MSE. We do not plot RCT because the algorithm obtained a median MSE of 37.29. RCT thus came in second place in STAR*D but performed terribly in this problem. These results are expected because most machine learning algorithms cannot consistently extrapolate well. We also find that SDD again outperforms its ablated variants. We conclude that SDD achieves the best performance when trained on adults but tested on children with the RCT data. 

\section{Conclusion} \label{sec_concl}
We introduced SDD to estimate the CATE across the entire clinical population. SDD first approximates pre- and post-treatment conditional expectations using the observational data. The algorithm then optimally combines them with an RCT, effectively relaxing the parallel slopes assumption used in the original DD formulation. In practice, SDD outperforms prior approaches by weighting the component conditional expectations used to estimate the CATT. SDD can thus help investigators generalize results from RCTs to the clinical population at large with a high degree of accuracy. 

\begin{acks}
We thank the anonymous reviewers for their time, insights and recommendations.
\end{acks}

\bibliographystyle{ACM-Reference-Format}
\bibliography{biblio}

\newpage
\appendix
\section{Proofs}

\begin{repproposition}{thm_psiCATT}
The CATT is equal to: 
\begin{equation} \nonumber
\begin{aligned}
    \psi_\beta(\bm{X}) \triangleq &\hspace{1mm}[\mathbb{E}(Y_1(1)|\bm{X},T=1) - \mathbb{E}(Y_0(1)|\bm{X},T=1)\beta_1]\\ &- [\mathbb{E}(Y_1(0)|\bm{X},T=0)\beta_2 - \mathbb{E}(Y_0(0)|\bm{X},T=0)\beta_3],
\end{aligned}
\end{equation}
under Assumptions \ref{as_preT} and \ref{as_NPslope}.
\end{repproposition}
\begin{proof}
We can write:
\begin{equation} \nonumber
\begin{aligned}
        &\mathbb{E}(Y_1(1) - Y_1(0)|\bm{X},T=1)\\
         \stackrel{\ref{as_preT}}{=} &\hspace{1mm}\mathbb{E}(Y_1(1)|\bm{X},T=1) - \mathbb{E}(Y_1(0)|\bm{X},T=1)\\ &+ \mathbb{E}(Y_0(0)|\bm{X},T=1)\beta_1  - \mathbb{E}(Y_0(1)|\bm{X},T=1) \beta_1\\
         = &\hspace{1mm}\mathbb{E}(Y_1(1)|\bm{X},T=1) - [\mathbb{E}(Y_1(0)|\bm{X},T=1)\\ &- \mathbb{E}(Y_0(0)|\bm{X},T=1)\beta_1 ] - \mathbb{E}(Y_0(1)|\bm{X},T=1) \beta_1\\
        \stackrel{\ref{as_NPslope}}{=} &\hspace{1mm}\mathbb{E}(Y_1(1)|\bm{X},T=1) - [\mathbb{E}(Y_1(0)|\bm{X},T=0)\beta_2 \\ &- \mathbb{E}(Y_0(0)|\bm{X},T=0)\beta_3] - \mathbb{E}(Y_0(1)|\bm{X},T=1) \beta_1\\
        = &\hspace{1mm}[\mathbb{E}(Y_1(1)|\bm{X},T=1) - \mathbb{E}(Y_0(1)|\bm{X},T=1)\beta_1]\\ &- [\mathbb{E}(Y_1(0)|\bm{X},T=0)\beta_2 - \mathbb{E}(Y_0(0)|\bm{X},T=0)\beta_3]\\
        = &\hspace{1mm}\psi_{\beta}(\bm{X}).
\end{aligned}
\end{equation}
\end{proof}

\begin{reptheorem}{thm:RCT}
Expression \eqref{eq:RCTreg} is equivalent to $\mathbb{E}(Y_1(1)|\bm{X},T=1) + H(\bm{X})\beta$ in the RCT distribution under Assumptions \ref{ax_RCT}, \ref{as_SB} and \ref{as_CATEslope}.
\end{reptheorem}
\begin{proof}
We may write the following sequence with the RCT distribution:
\begin{equation} \nonumber
\begin{aligned}
    &\mathbb{E}(Y_1(1)| \bm{X},T=1,\bm{S}=1) - \mathbb{E}(Y_1(0)| \bm{X},T=0,\bm{S}=1)\\
        \stackrel{\ref{ax_RCT}}{=} \hspace{1mm} &\mathbb{E}(Y_1(1) - Y_1(0) | \bm{X},\bm{S}=1)\\
        \stackrel{\ref{as_SB}}{=}\hspace{1mm} &f(\bm{X})\\
        \stackrel{\ref{as_CATEslope}}{=}\hspace{1mm} &[\mathbb{E}(Y_1(1)|\bm{X},T=1) - \mathbb{E}(Y_0(1)|\bm{X},T=1)\beta_1]\\ &- [\mathbb{E}(Y_1(0)|\bm{X},T=0)\beta_2 - \mathbb{E}(Y_0(0)|\bm{X},T=0)\beta_3]\\
        = &\hspace{1mm}\mathbb{E}(Y_1(1)|\bm{X},T=1) + H(\bm{X})\beta.
\end{aligned}
\end{equation}
\end{proof}

\begin{reptheorem}{thm:fisher} (Fisher consistency)
If $\mathbb{E}(H(\bm{X})^TH(\bm{X}))$ is invertible, then the population version of SDD recovers $f(\bm{X})$ for any $\bm{X} \in \mathcal{S}_O$ under Assumptions \ref{as_support}-\ref{as_SB} and \ref{as_CATEslope}.
\end{reptheorem}
\begin{proof}
 Assumptions \ref{ax_RCT} and \ref{as_SB} ensure that Expression \eqref{eq:RCTreg} is equivalent to $f(\bm{X})$ for any $\bm{X} \in \mathcal{S}_R$ for Line \ref{alg:iT}. We know that $\mathbb{E}(Y_1(1) | \bm{X},T=1)$ and the components of $H(\bm{X})$ are unique for any $\bm{X} \in \mathcal{S}_O$ for Line \ref{alg:G}. The conclusion follows because $\beta$ is the unique minimizer of the expected squared L2-norm between $k(\bm{X})$ and $H(\bm{X})$ for any $\bm{X} \in \mathcal{S}_R$ in Line \ref{alg:eT} by Assumption \ref{as_support}, Theorem \ref{thm:RCT} and the invertibility of $\mathbb{E}(H(\bm{X})^TH(\bm{X}))$. Hence, SDD recovers $\mathbb{E}(Y_1(1)|\bm{X},T=1) + H(\bm{X})\beta$ for any $\bm{X} \in \mathcal{S}_O$.
\end{proof}

We use the notation $\bm{X}_n = O_p(r(n))$ for a potentially multivariate random vector $\bm{X}_n$ if $\|\bm{X}_n \| = O_p(r(n))$. Observe that $\bm{X}_n = O_p(r(n))$ if and only if each entry of $\bm{X}_n$ lies in $O_p(r(n))$. The above statements hold likewise for $o_p(1)$.

We use the notation $\underline{H}$ to refer to a matrix containing $2m$ row-samples of $H(\bm{X})$, and likewise for other random variables.
\begin{lemma1} \label{lem_inv}
If $\mathbb{E}(H(\bm{X})^TH(\bm{X}))$ is finite and invertible, then\\ $ \Big(\frac{1}{2m} \underline{H}^T \underline{H} \Big)^{-1} = O_p(1)$. Further if $\widehat{H}(\bm{X}) - H(\bm{X}) = o_p(1)$, then $ \Big(\frac{1}{2m} \underline{\widehat{H}}^T \underline{\widehat{H}} \Big)^{-1} = O_p(1)$.
\end{lemma1}
\begin{proof}
We know that $A \triangleq \frac{1}{2m} \underline{H}^T \underline{H} - \mathbb{E}(H(\bm{X})^TH(\bm{X})) = o_p(1)$ by the central limit theorem, since $H(\bm{X})$ has finite covariance. The first conclusion follows by the continuous mapping theorem; we have $\Big(\frac{1}{2m} \underline{H}^T \underline{H}\Big)^{-1} - \mathbb{E}^{-1}(H(\bm{X})^TH(\bm{X})) = o_p(1)$. For the second, we know that  $B \triangleq \frac{1}{2m} \underline{\widehat{H}}^T \underline{\widehat{H}} - \frac{1}{2m} \underline{H}^T \underline{H} = o_p(1)$ also by the continuous mapping theorem. We thus have $A + B = o_p(1)$, so the second conclusion follows by applying the continuous mapping theorem again for the outer inversion.
\end{proof}

\begin{reptheorem}{thm_conv} (Convergence rate)
Consider Assumptions \ref{as_support}-\ref{as_SB} and \ref{as_CATEslope}. Further assume:
\begin{enumerate}
\item The regression procedures in Lines \ref{alg:iT} and \ref{alg:G} converge to their estimands at rate $O_p(s(m)) = o_p(1)$ on $\mathcal{S}_R$ and $O_p(r(n)) = o_p(1)$ on $\mathcal{S}_O$, respectively.
\item $H(\bm{X})^Tk(\bm{X})$ has finite covariance.
\item $\mathbb{E}(H(\bm{X})^TH(\bm{X}))$ is finite and invertible. 
\end{enumerate}
Then SDD approximates the true CATE for any $\bm{X} \in \mathcal{S}_O$ at rate $O_p(s(m)) + O_p(r(n)) + O_p(1/\sqrt{m})$ corresponding to Lines \ref{alg:iT}, \ref{alg:G} and \ref{alg:eT}, respectively.
\end{reptheorem}
\begin{proof}
We suppress most functional inputs in parentheses to remove notational clutter. Consider the column vectors of the RCT data: $\gamma_{Hk} = [1,\mathbb{E}(H^TH)^{-1}\mathbb{E}(H^Tk)]^T$, $\widehat{\gamma}_{Hk} = [1,(\frac{1}{2m} \underline{\widehat{H}}^T \underline{\widehat{H}})^{-1}\\(\frac{1}{2m} \underline{H}^T \underline{k})]^T$, and likewise for $\widehat{\gamma}_{\widehat{H}k}$ and $\widehat{\gamma}_{\widehat{H}\widehat{k}}$. Let $L$ refer to the row vector $[\mathbb{E}(Y_1(1)|\bm{X},T=1),H]$.

It suffices to show that $\| L \gamma_{Hk} - \widehat{L}\widehat{\gamma}_{\widehat{H}\widehat{k}}\| = O_p(r(n)) + O_p(1/\sqrt{m}) + O_p(s(m))$ by Theorem \ref{thm:RCT}. We write the following sequence of inequalities:
\begin{equation} \nonumber
\begin{aligned}
    &\| L \gamma_{Hk} - \widehat{L}\widehat{\gamma}_{\widehat{H}\widehat{k}}\|\\
    \leq \hspace{1mm}& \| L \gamma_{Hk} - \widehat{L}\gamma_{Hk}\| + \|\widehat{L} \gamma_{Hk} - \widehat{L} \widehat{\gamma}_{Hk}\|\\ \hspace{1mm}& +  \|\widehat{L} \widehat{\gamma}_{Hk} - \widehat{L} \widehat{\gamma}_{\widehat{H}k}\| + \|\widehat{L} \widehat{\gamma}_{\widehat{H}k} - \widehat{L} \widehat{\gamma}_{\widehat{H}\widehat{k}}\|\\
    = \hspace{1mm}& \underbrace{\|L\gamma_{Hk} - \widehat{L}\gamma_{Hk}\|}_A +  \\
    &\hspace{2mm}\Big(\underbrace{\Big\| \widehat{L}\mathbb{E}^{-1}(H^T H) \mathbb{E}(H^Tk) - \widehat{L}\Big(\frac{1}{2m} \underline{\widehat{H}}^T \underline{\widehat{H}}\Big)^{-1} \frac{1}{2m} \underline{H}^T \underline{k} \Big\|}_B\\
    &\hspace{2mm}+ \underbrace{\Big\| \widehat{L}\Big(\frac{1}{2m} \underline{\widehat{H}}^T \underline{\widehat{H}}\Big)^{-1} \frac{1}{2m} \underline{H}^T \underline{k} - \widehat{L}\Big(\frac{1}{2m} \underline{\widehat{H}}^T \underline{\widehat{H}}\Big)^{-1} \frac{1}{2m} \underline{\widehat{H}}^T \underline{k} \Big\|}_C\\ 
    &\hspace{2mm}+ \underbrace{\Big\|\widehat{L}\Big(\frac{1}{2m} \underline{\widehat{H}}^T \underline{\widehat{H}}\Big)^{-1} \frac{1}{2m} \underline{\widehat{H}}^T \underline{k} - \widehat{L}\Big(\frac{1}{2m} \underline{\widehat{H}}^T \underline{\widehat{H}}\Big)^{-1}\frac{1}{2m} \underline{\widehat{H}}^T\underline{\widehat{k}}\Big\|}_D \Big)
\end{aligned}
\end{equation}
$A$ is in $O_p(r(n))$ for any $\bm{X} \in \mathcal{S}_O$ by assumption. Note that $\widehat{L} = O_p(1)$ for $B$, $C$ and $D$. $B$ is in $O_p(1/\sqrt{m})$ for any $\bm{X} \in \mathcal{S}_O$ by invoking Slutsky's theorem after noting that (1) $\Big(\frac{1}{2m} \underline{\widehat{H}}^T \underline{\widehat{H}} \Big)^{-1} \stackrel{p}{\rightarrow} \mathbb{E}^{-1}(H^T H)$ by the continuous mapping theorem with Assumption \ref{as_support}, and (2) $\sqrt{2m}\Big(\frac{1}{2m} \underline{H}^T \underline{k} - \mathbb{E}( H^T k) \Big)$ converges to a normal distribution by the central limit theorem due to the finite covariance of $H^Tk$. 

For $C$, note that $\Big(\frac{1}{2m} \underline{\widehat{H}}^T \underline{\widehat{H}}\Big)^{-1} = O_p(1)$ by Assumption \ref{as_support} and Lemma \ref{lem_inv}. Hence, $C$ lies in $O_p(1)O_p(r(n)) = O_p(r(n))$ for any $\bm{X} \in \mathcal{S}_O$. 

For $D$, we again use $\Big(\frac{1}{2m} \underline{\widehat{H}}^T \underline{\widehat{H}}\Big)^{-1} = O_p(1)$. Let $g = \mathbb{E}(Y_1(1)|\bm{X},T=1)$. We have:
\begin{equation} \nonumber
\begin{aligned}
   &\Big\| \frac{1}{2m} \underline{\widehat{H}}^T \underline{k} - \frac{1}{2m} \underline{\widehat{H}}^T\underline{\widehat{k}}\Big\| \\
   \leq \hspace{2mm}&\Big\| \frac{1}{2m} \underline{\widehat{H}}^T \underline{f} - \frac{1}{2m} \underline{\widehat{H}}^T\underline{\widehat{f}}\Big\| + \Big\| \frac{1}{2m} \underline{\widehat{H}}^T \underline{g} - \frac{1}{2m} \underline{\widehat{H}}^T\underline{\widehat{g}}\Big\|
\end{aligned}
\end{equation}
Further recall that $\widehat{H} = O_p(1)$. $D$ is thus $O_p(1)O_p(r(n)) + O_p(1) O_p(s(m))\\ =O_p(r(n)) + O_p(s(m))$ for any $\bm{X} \in \mathcal{S}_O$. The conclusion follows because $A + B + C + D = 3O_p(r(n)) + O_p(1/\sqrt{m}) + O_p(s(m)) = O_p(r(n)) + O_p(1/\sqrt{m}) + O_p(s(m))$.
\end{proof}
\end{document}